\documentclass[letterpaper, 10 pt, conference]{ieeeconf}  

\IEEEoverridecommandlockouts                              

\usepackage{graphicx} 
\usepackage{epsfig} 
\usepackage{mathptmx} 
\usepackage{times} 
\usepackage{amsmath} 
\usepackage{amssymb}  
\usepackage{multirow}
\usepackage{booktabs}
\usepackage{graphics}
\usepackage{tabularx}
\usepackage{makecell}
\usepackage{stackengine}
\usepackage{xcolor}
\usepackage{colortbl}
\usepackage{array}
\usepackage{bm}
\usepackage{amsmath}
\usepackage{hyperref} 
\usepackage[capitalize]{cleveref}
\usepackage{cite}
\usepackage{threeparttable}
\usepackage{diagbox}
\usepackage{subfig}
\usepackage{xspace}

\newcommand{\etal}{\emph{et al.}\@\xspace}

\newcommand{\eg}{\emph{e.g.}\@\xspace}

\title{\LARGE \bf
Co-MTP: A Cooperative Trajectory Prediction Framework \\ with Multi-Temporal Fusion for Autonomous Driving
}

\author{Xinyu Zhang$^{1*}$, Zewei Zhou$^{1*}$, Zhaoyi Wang$^{1}$, Yangjie Ji$^{1}$, Yanjun Huang$^{1}$, Hong Chen$^{1}$ 
\thanks{$^{*}$Equal contribution.}
\thanks{$^*$Supported by the National Natural Science Foundation of China, Joint Fund for Innovative Enterprise Development (U23B2061)}
\thanks{
 $^{1}$School of Automotive Studies, Tongji University, Shanghai, China. {\tt\small\{2332922, zhouzewei, 2233591, jiyj20, yanjun\_huang, chenhong2019\}@tongji.edu.cn}}%
}

\begin{document}

\maketitle
\thispagestyle{empty}
\pagestyle{empty}

\begin{abstract}


Vehicle-to-everything technologies (V2X) have become an ideal paradigm to extend the perception range and see through the occlusion. Exiting efforts focus on single-frame cooperative perception, however, how to capture the temporal cue between frames with V2X to facilitate the prediction task even the planning task is still underexplored. In this paper, we introduce the Co-MTP, a general cooperative trajectory prediction framework with multi-temporal fusion for autonomous driving, which leverages the V2X system to fully capture the interaction among agents in both history and future domains to benefit the planning. In the history domain, V2X can complement the incomplete history trajectory in single-vehicle perception, and we design a heterogeneous graph transformer to learn the fusion of the history feature from multiple agents and capture the history interaction. Moreover, the goal of prediction is to support future planning. Thus, in the future domain, V2X can provide the prediction results of surrounding objects, and we further extend the graph transformer to capture the future interaction among the ego planning and the other vehicles' intentions and obtain the final future scenario state under a certain planning action. We evaluate the Co-MTP framework on the real-world dataset V2X-Seq, and the results show that Co-MTP achieves state-of-the-art performance and that both history and future fusion can greatly benefit prediction. Our code is available on our project website:
\href{https://xiaomiaozhang.github.io/Co-MTP/}{https://xiaomiaozhang.github.io/Co-MTP/}

\end{abstract}

\section{INTRODUCTION}

Autonomous vehicles (AVs) should interact with complex and dynamic environments to ensure safe and efficient operation \cite{hu2023planning, yuan2024evolutionary, zhou2022comprehensive}. Thus, the trajectory prediction has attracted the community's attention \cite{huang2022survey, shi2024mtr++, ngiam2021scene}. The accurate prediction relies on the comprehensive perception information, especially in complex and dense scenarios \cite{lange2024scene}. The limited perception range and occlusion always lead to catastrophic results \cite{han2024foundation,yangjie2024towards,ji2023optimization}, however, most existing works assume that all surrounding objects are fully visible and have complete history trajectories \cite{jia2023hdgt, zeng2021lanercnn, zhao2021tnt}, which is unreal in real vehicle onboard sensors. Thus, vehicle-to-everything (V2X) technology has been an ideal paradigm for sharing information among agents and seeing through occlusion. Nevertheless, existing V2X efforts focus on the single-frame cooperative perception \cite{xu2022opv2v, xiang2024v2x, xu2023v2v4real}. Issues such as \textit{How to capture the temporal cue between frames with V2X to facilitate the prediction task and even the planning task, and how to fully employ the V2X information in real-world environments with occlusion} are still underexplored and present deployment challenges.

In the real world, each agent's sensor differs in capability, accuracy, and coverage \cite{cai2023analyzing, ruan2023learning}, and the positioning of these sensors can lead to varied levels of perception and occlusion of the same object. The simplest method of trajectory fusion is association and stitching \cite{yu2023v2x}, but it cannot handle the diverse errors from different sensors and still faces the incomplete problem when multiple agents lose the same timestamp in dense scenarios. Thus, it is necessary to explore how cooperative prediction integrates and leverages incomplete history trajectories from multiple agents to obtain more comprehensive history information. 



Furthermore, the goal of prediction is not merely to predict the motion of surrounding objects but to support the ego vehicle’s planning \cite{hu2023planning, huang2023differentiable}. Each planning candidate represents a driving intent, and the vehicle should evaluate each candidate by reasoning the future because the reaction of the environment varies depending on the ego's actions. If the AV reasons the surroundings solely based on the isolated prediction, it will become too afraid to drive and will always consistently yield to other road users \cite{liu2021deep, song2020pip}. On the contrary, proactive planning and prediction make AV consider their intention first and the reason for the other reaction. However, the ego planning trajectory is the only future feature in the future domain, leading to aggressive driving behaviors and danger. Thus, for planning-oriented prediction, it is necessary not only to incorporate history interactions but also to consider future interactions among surrounding objects.

To address these challenges, we present Co-MTP, a \underline{\textbf{Co}}operative \underline{\textbf{M}}ulti-\underline{\textbf{T}}emporal fusion \underline{\textbf{P}}rediction framework. First, the infrastructure, with elevated and stationary perspectives, can provide perception and prediction support to each agent in the scenario, \eg AV, human driver, and pedestrian. Second, when AV drives into the infrastructure area, it can receive the infrastructure's history and prediction information, and fuse them in both history and future domains with the planning information of the ego vehicle to facilitate the planning-oriented prediction. Our contributions can be summarized as follows:

\begin{figure*}[t]
    \centering
    \includegraphics[width=0.94\linewidth]{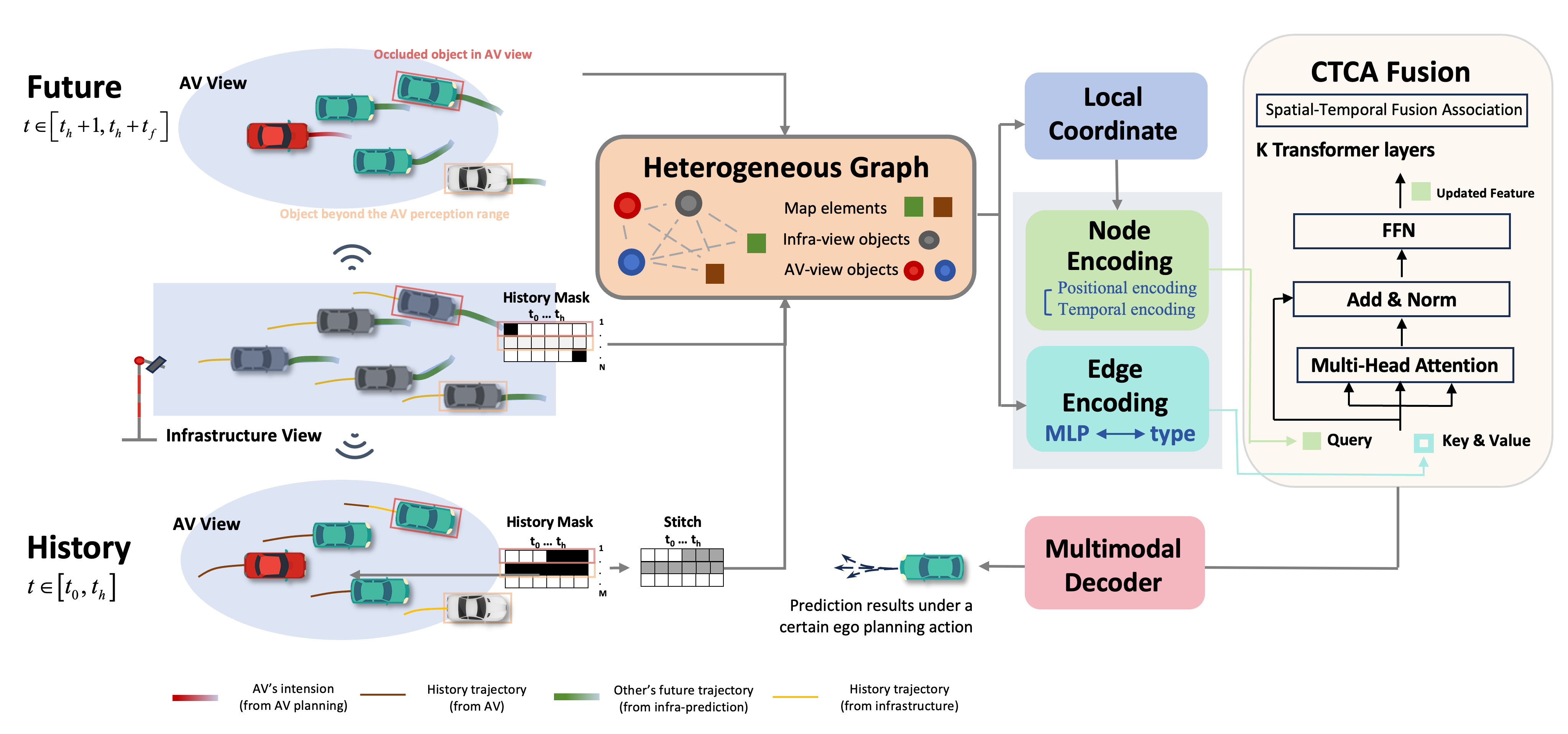}
    \caption{\textbf{The overall architecture of Co-MTP.}
    In this framework, infrastructures share the history and their prediction results to ego AV. Then, we construct a heterogeneous scene graph with the processed trajectory data and map information, categorizing them according to the types of objects and map elements. Next, we initialize the features of nodes and edges in the relative coordinate system of each object. The CTCA Fusion is used to update the features of the nodes and edges selected by the STSA module over K Transformer layers. Finally, we take the nodes' hidden features from the last layer and input them into the Multimodal Decoder to obtain the multimodal trajectory prediction results.}
    \label{fig: Co-MTP}
    \vspace{-0.6cm}
\end{figure*}

\begin{itemize}
\item We propose Co-MTP, a general cooperative trajectory prediction framework with multi-temporal fusion across both history and future domains. Co-MTP is the first framework to fully fuse and exploit comprehensive temporal information in prediction with V2X.
\item For the issue of incomplete trajectory in the history domain, we designed a heterogeneous graph to learn the fusion of the history feature from multiple agents with transformers.
\item To facilitate the following planning, we further extended the heterogeneous graph to future domains to capture the future interaction among the ego planning and the other vehicles' actions from infrastructure prediction. 
\item We evaluate the Co-MTP framework on real-world dataset V2X-Seq. The results demonstrate its state-of-the-art performance, and both history and future fusion can greatly benefit prediction and planning.


\end{itemize}


\section{RELATED WORK}

\noindent \textbf{Cooperative Prediction.} The existing effort of V2X research focuses on cooperative perception, and cooperative prediction which considers the temporal information with V2X is still in its infancy \cite{wu2024cmp, araluce2024enhancing, zhou2024v2xpnp}. V2VNet \cite{vedaldi_v2vnet_2020} is the first work to explore this field, which proposed an end-to-end perception and prediction model. However, its prediction head is simple and ignores the map information and complex interaction features. Yu, \etal \cite{yu2023v2x} presented a open-source V2X sequence dataset for cooperative prediction, V2X-Seq. V2XPnP \cite{zhou2024v2xpnp} provides an open-source general framework and V2X sequential dataset for end-to-end cooperative perception and prediction. Furthermore, the V2X-Graph \cite{ruan2023learning} is proposed based on V2X-Seq, which provides a graph neural network (GNN) to learn the cooperative trajectory representations. However, how to fully employ the V2X temporal information, such as the different temporal domains, and fuse the cooperative trajectory still needs to be explored.

\noindent \textbf{Planning-informed Prediction.} The couple of prediction and planning is the focus for safe driving in interaction scenarios \cite{hu2023planning, espinoza2022deep}. PiP model \cite{song2020pip} directly incorporated the planning information into prediction, and Liu, \etal \cite{liu2021deep} and Huang, \etal \cite{huang2023differentiable}, further merge the prediction module and general optimization-based  planning module \cite{bhardwaj2020differentiable, zhou2021reliable} to construct a reaction framework. However, only considering ego planning in the future domain will inadvertently overestimate its priority on the road, leading to overly optimistic predictions and potentially hazardous behaviors \cite{tang2022interventional}. Therefore, it is crucial to integrate the future intentions of other vehicles during prediction to ensure a comprehensive extraction of interaction features across both history and future domains. The other's intention is prediction targets in single-vehicle prediction, and the ego vehicle not only can access history information to see through occlusion, but also the future information as the prior knowledge.

\noindent \textbf{Interaction-aware Prediction.} Trajectory prediction is one of the most critical research areas in autonomous driving \cite{huang2022survey}, which considers the perception results sequence and map information to capture the interaction context and predict the others' trajectories \cite{gao2020vectornet}. The interaction feature extractor has developd from CNN \cite{deo2018convolutional, song2020pip} to GNN and Transformer \cite{zhou2022hivt, zhou2022comprehensive}. The Transformer uses multi-attention to capture the temporal feature across a sequence without relying on sequential order processing \cite{shi2024mtr++,zhou2023qcnext}, which is an advantage in handling incomplete trajectories and fusing the trajectory feature from different agents. Moreover, GNN builds the complex interaction context as a graph with nodes and edges \cite{trajectron++,jia2023hdgt}, which can facilitate the learning of complex interaction, especially in cooperative prediction involving multi-agent and multi-temporal features.

\section{Methodology}


Fig. \ref{fig: Co-MTP} illustrates the Co-MTP framework. Both AVs and infrastructures can operate prediction individually to support the AV and scenario object separately. The input of the vehicle-side prediction model is not limited to the history domain but also includes the prediction results from the infrastructure-side prediction model in the future domain.

\subsection{Problem Definition} 

The input comprises two parts: history trajectories, AV's future ground truth and map data. Consider a scenario where multiple viewpoints jointly observe a set of $Z$ objects $\Omega=\{a_i | i=1, 2, ..., Z\}$ from time step $t_0$ to $t_p$. Among them, the AV perceives $M$ objects $\Omega_{AV}=\{b_i | i=1, 2, ..., M\}$ and $b_i \in \Omega$, while the infrastructure perceives $N$ agents $\Omega_{infra}=\{c_i | i=1, 2, ..., N\}$ and $c_i \in \Omega$. All objects' features are given, including their bounding boxes (length, width, height) and types (pedestrian, bicycle, vehicle). Define the object trajectory $i$ as $X_i=\left\{\bm{x}^{t_0}_i, \bm{x}^{t_1}_i, ..., \bm{x}^{t_h}_i\right\}$ in a certain frequency, where $\bm{x}^{t}_i$ includes coordinates, heading, velocity, etc. Due to occlusion and other reasons, the observation could be missing in some frames. Thus, each object's trajectory has a corresponding mask vector $\bm{m}_i=\left\{m^{t_0}_i, m^{t_1}_i, ..., m^{t_h}_i\right\}$, which indicates the data validity. If the data exists at time step $t$, then $m^t_i=1$; otherwise, the value is $0$. In addition, to simulate the planning trajectory from the real-world dataset, we follow the setting of \cite{song2020pip}, and use the quintic curve to interpolate the future ground truth to follow the planning mission goal. The map data includes roads and signals, and all roads are represented as polylines or polygons, such as road lines, crosswalks, and stop-lines. 


The prediction goal is to use the aforementioned information to predict the future trajectories of the surrounding objects. In other words, the task is to learn a model $f(.)$ that outputs $Y_i=\left\{\bm{y}^{t_h+1}_i, \bm{y}^{t_h+2}_i, ..., \bm{y}^{t_h+t_f}_i\right\}$, Additionally, to capture the uncertainty, the predicted output could be multiple modes, each associated with a corresponding probability.

\subsection{Scene Representing with Graph}

\noindent \textbf{Heterogeneous Graph.} The driving scenario has a lot of heterogeneous elements, such as map, trajectories, different object types and traffic signals. To leverage the powerful scenario-representation capabilities, we build a heterogeneous graph for prediction scenario $\mathcal{G} = \left\{\mathcal{N}, \mathcal{E}\right\}$, where $\mathcal{N}$ and $\mathcal{E}$ present the set of nodes and edges separately. Objects and the map are represented as graph nodes, and the relationships between them are represented as edges between the nodes. To represent heterogeneity, we construct a type mapping function $\mathcal{T}$ for the nodes and define the type of each edge based on its originating node.

\noindent \textbf{Trajectory Data Preprocess.} In Co-MTP framework, objects' history trajectories come from both the AV and infrastructure perspectives, along with the corresponding mask matrices. Before learning and fusing the information from these perspectives, we follow the stitch method CBMOT \cite{benbarka2021score} in V2X-Seq \cite{yu2023v2x} to use the high-detection-score to interpolate the vehicle history trajectory with infrastructure history, and the masks also are reorganized. We found the prestitch of the incomplete trajectory can complement the temporal index and accelerate the learning of the following trajectory fusion. 
This prestitch method does not mean that the infrastructure information not involved in the interpolation is discarded. All the original objects' history trajectories from the infrastructure are encoded in the graph nodes.
In the future domain, in addition to the AV's planning trajectory, the infrastructure's prediction results of the other objects are also included. Since our research does not focus on the infrastructure's prediction task itself, we use the our Co-MTP model with only infrastructure information to output other agents' predicted trajectories.

To reduce the computational burden, the number of edges for each node has the upper boundary. Whether the node $\bm{n}$ can establish an edge with the node $\bm{s}$ depends on the distance between them. For object nodes, we use the position observed at the last time step $t_h$ to calculate the distance to others. For map nodes, the shortest distance from other nodes to the polyline or polygon is selected.

\subsection{Relative Spatial-Temporal Encoding} 

\noindent \textbf{Local Coordinate System.} To facilitate interaction learning in a normalized form, each node requires a local coordinate system $(x_{\rm{ref}}, y_{\rm{ref}}, \theta_{\rm{ref}})$ to provide a unique motion representation for each object, regardless of its global coordinates. All subsequent computations related to the node will be carried out within this local coordinate system. The position and heading of each object at the final time step $t_h$ are taken as $(x_{\rm{ref}}, y_{\rm{ref}})$ and $\theta_{\rm{ref}}$. For map data, the center point coordinates are used as $(x_{\rm{ref}}, y_{\rm{ref}})$, and the direction of the longest line segment within the polyline or polygon is treated as $\theta_{\rm{ref}}$.



\noindent \textbf{Graph Node Encoding.} All the positions of objects from both AV and infrastructure are encoded with a shared-parameter MLP. Then, the position features are concatenated with other information (\eg bounding box and speed) input into two distinct MLPs—one designed for the AV and the other for the infrastructure with different parameters. We employ two self-attention layers to capture the temporal features of each object in both history and future domains. Since the history trajectory may be incomplete, the mask vector $\bm{m}_i$ is also input to the temporal encoding. 
For map data, we combine the MLP used for coordinate encoding with an improved PointNet to form the map position encoding method. Additionally, we use a set of Linear layers and GRUs to encode traffic signals. 
Thus, all nodes have been encoded into a set $\left\{\bm{n}^0\right\}$.

\noindent \textbf{Graph Edge Encoding.} All edges are first input into a shared-parameter MLP to encode the coordinate transformation $\Gamma_{\bm{s} \rightarrow \bm{n}}$. According to the type mapping function $f(.)$, all edges are connected to their starting nodes and the combined input is fed into the corresponding type-specific MLP to generate the initial edge features.
\begin{equation}
    \bm{e}^{0}_{\bm{s} \rightarrow \bm{n}} = \mathrm{MLP}_{\bm{s}}\left( \mathrm{concat} \left[\bm{s}^{0},  \mathrm{MLP}\left(\Gamma_{\bm{s} \rightarrow \bm{n}}  \right) \right] \right)
\end{equation}

\subsection{Cross-Temporal and Cross-Agent (CTCA) Fusion}

To capture the complex trajectory features from multiple agents, we propose a cross-temporal graph with the spatial-temporal fusion association (STFA) module. A K-layer Transformer is employed to aggregate and update features for nodes and edges.

\begin{figure}
    \centering
    \includegraphics[width=\linewidth]{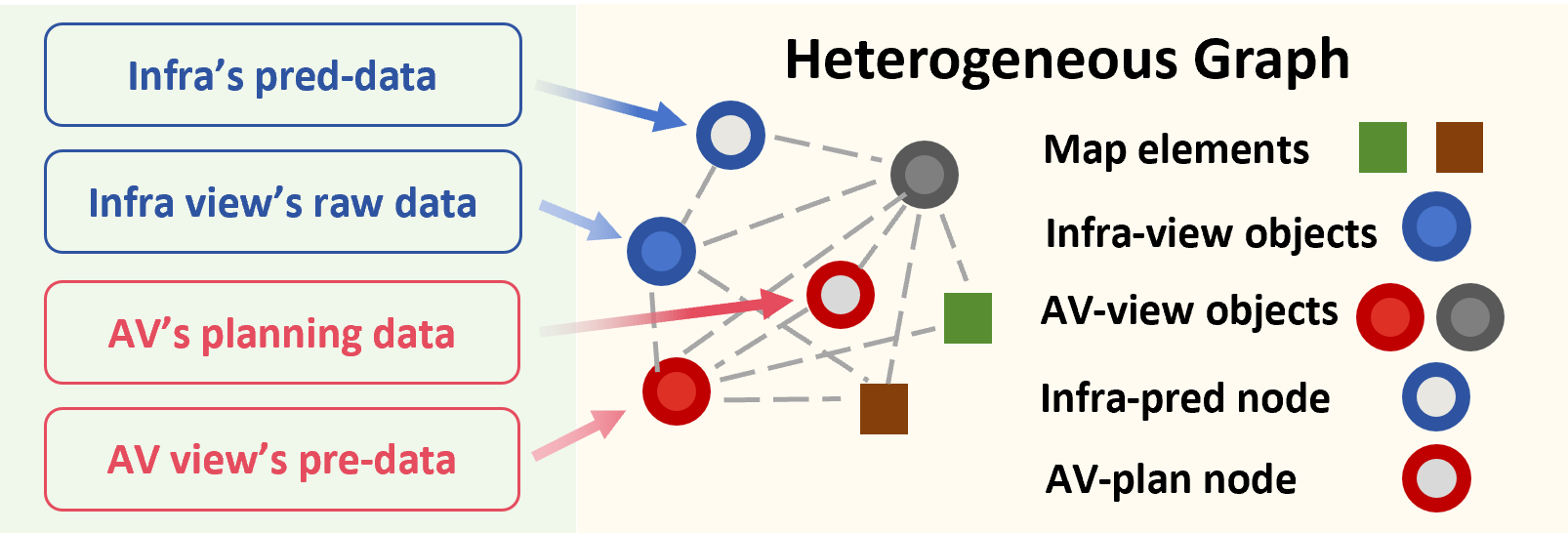}
    \caption{Illustration of STFA in the heterogeneous graph. In addition to the object nodes with preprocessed data from the AV's view, objects with raw data from the infrastructure's view also participate as independent nodes. To model future interaction, we treat the AV'planning and infrastructure's prediction results as independent nodes, establishing edge relationships with historical nodes in the graph.}
    \label{fig: STFA}
    \vspace{-0.6cm}
\end{figure}

\noindent \textbf{Spacial-Temporal Fusion Association.} Fig. \ref{fig: STFA} illustrates the STFA module. It aims to fully integrate information from different views and temporal domains. AV and infrastructure can observe the same object in the same timestamp and have separate history information record, thus, it is necessary to establish a connection between these two agents' information. We employ an implicit approach to extract the multi-view features across multi-temporal domain.
When gathers information, the vehicle-side node $\bm{v}$, after trajectory interpolation, is continuously influenced by the infrastructure node $\bm{i}$. 
This not only reuses the infrastructure information that was not utilized during the interpolation but also learn the relationship between the two views. Furthermore, when aggregating agent node, future information should be considered. Similarly to the above approach, future nodes provide the agents with the necessary information for aggregation.

\noindent \textbf{Transformer Layer.} We utilize the multi-head attention layer (MHA) to capture the feature from various types of in-edges for each node, followed by an MLP to combine the information. During the aggregation phase, the node features are calculated as the query ($\mathbf{Q}$), and the in-edge features are used as the key ($\mathbf{K}$) and the value ($\mathbf{V}$). 
\begin{equation}
    \bm{n}^{k-1,'}_{type_e} = 
    \mathrm{MHA}_{type_e}\left(\bm{n}^{k-1}, \bm{e}^{k-1}_{\bm{s} \rightarrow \bm{n}}\right)
\end{equation}
\begin{equation}
    \bm{n}^{k-1,''} = \mathrm{MLP}\left(
    \mathrm{concat}\left(
    \left\{\bm{n}^{k-1,'}_{type_e} \mid type_e = 1, ..., n_r\right\}
    \right)
    \right)
\end{equation}
where $type_e$ means the type of $\bm{e}^{k-1}_{\bm{s} \rightarrow \bm{n}}$, and $n_r$ is number of the in-edges' types for node $\bm{n}$.
After that, the features of each type of node are updated with a feed-forward network (FFN).
\begin{equation}
    \bm{n}^k = \mathrm{FFN}_{\bm{n}}\left(\bm{n}^{k-1,''}
    \right)
\end{equation}
where $\bm{n}^k$ is the hidden feature of the $k^{th}$ Transformer. $\mathrm{FFN}_{\bm{n}}$ means the parameters depend on the type of node $\bm{n}$.

For edge feature aggregation and updating, we retrieve the outputs $\bm{s}^{k-1}$ and $\Gamma_{\bm{s} \rightarrow \bm{n}}$ from the edge encoding stage and combine them with the edge features from the previous layer, $\bm{e}^{k-1}_{\bm{s} \rightarrow \bm{n}}$. These combined features are then input into an MLP to obtain the updated edge features.

\begin{equation}
    \bm{e}^{k}_{\bm{s} \rightarrow \bm{n}}=\mathrm{MLP}_{\bm{s} \rightarrow \bm{n}}
    \left( \mathrm{concat} 
    \left[\bm{s}^{k-1}, \Gamma_{\bm{s} \rightarrow \bm{n}}, \bm{e}^{k-1}_{\bm{s} \rightarrow \bm{n}} 
    \right] 
    \right)
\end{equation}

\subsection{Multimodal Decoder}

The motion pattern of an agent is inherently multi-modal, where multiple future trajectory possibilities coexist.The hidden features output by the final layer of the transformer layer typically consist of a single set of feature embeddings.
In this paper, we first replicate these single embeddings and their corresponding in-edge features $K$ times. Then, we refine the node information through a cross-attention mechanism with processed in-edge features to extract richer feature representations and output $K$ modes via a feed-forward network (FFN).

Traditionally, the $K$ trajectory modes are treated as independent of one another. However, the modes of an object exhibit a competitive relationship, where the rise of one mode leads to the decline of others. Therefore, we input the refined features into a self-attention layer to capture the relationships between the $K$ modes of the same object.

After this refinement process, we employ both regression and classification techniques to decode the agent's trajectory. Specifically, a multi-layer perceptron (MLP) classifies the $K$ modalities, assigning each a probability. However, simply classifying the trajectory can lead to issues such as trajectory drift, which may violate the agent’s history tend. To mitigate this, we reintroduce the original node features $\bm{n}^0$ and impose dynamic constraints on the regression results. These refined features are concatenated with the regression output and passed through an MLP, with temporal aspects handled by a series of refined convolutional neural networks (CNNs). For a single agent, the final predicted output trajectory can be expressed as follows:
\begin{equation}
    Y_{cls} = \left\{
    c_k
    \right\}, k\in{\left[1, K\right]}
\end{equation}
\begin{equation}
    Y_{reg} = \left\{
    (\bm{y}^{1}_k, \bm{y}^{2}_k, ..., \bm{y}^{T}_k)
    \right\}, k\in{\left[1, K\right]}
\end{equation}
where $c_k$ is probability of the $k-th$ mode, $\bm{y}^{t}_k$ is the position of the agent at the timestamp $t$, the shape of tensor $Y_{cls}$ and $Y_{reg}$ are $(N\times{K})$ and $(N\times{K}\times{T}\times{2})$. $N$ is the number of target agents, $2$ means the 2-D coordinate $(x, y)$.

During the training, the total loss is calculated by the regression head and the classification head.
\begin{equation}
\begin{split}
    \mathcal{L} = & \lambda_1\mathcal{L}_{cls} + \lambda_2\mathcal{L}_{reg} \\    
    = &  \lambda_1\frac{1}{N}\sum_{n=1}^{N}\mathrm{CrossEntropy}
    \left(\left\{c_k\right\}_n,I(k_n)\right) \\
    &  +\lambda_2\frac{1}{2NT}\sum^N_{n=1}\sum^T_{t=1}
    \left[l(x^t_{n,k},x^{t*}_{n}) + l(y^t_{n,k},y^{t*}_{n})
    \right]
\end{split}
\end{equation}
where $\lambda_1$ and $ \lambda_2$ are the weights, which are set to 0.1 and 10. $I(k_n)$ is a hot vector, where the $k_n$-th element is 1 and others are 0. $k_n$ is the mode with the smallest regression loss, calculated as the average deviation between the regression results and the ground truth:
\begin{equation}
    k_n = \mathop{\mathrm{argmin}} \limits_{k\in{[1,K]}}\frac{1}{2T}\sum_{t=1}^{T}
    \left[l(x^t_{n,k},x^{t*}_{n}) + l(y^t_{n,k},y^{t*}_{n})
    \right]
\end{equation}
where $l(.)$ is smooth L1 loss.

\section{EXPERIMENTS}

\subsection{Experimental Setup}

\noindent  \textbf{Dataset and Metrics.} We evaluate the Co-MTP framework on the real-world V2X dataset, V2X-Seq, which comprises over 50,000 trajectory fragments of vehicle-infrastructure cooperation. Each fragment lasts for 10s, with a recorded frequency of 10 Hz. The dataset labels the type (\eg vehicle, bicycle, pedestrian, etc.) of each agent. To increase the sample size, we slice each original dataset fragment into 20 samples. Each sample contains 3s of history trajectories and 5s of future trajectories. 
According to standard evaluation protocol, we utilize metrics including minimum Average Displacement Error (minADE), minimum Final Displacement Error (minFDE), and Missing Rate (MR) for evaluation.

\noindent \textbf{Implementation Details.} All the trajectories are downsampled to 5Hz to accelerate the training. The map polylines and polygons are segmented, with each segment represented by 21 points. We select a map within 250m of the evaluation area and limit the number of edges for each node in the GNN to a maximum of 32. The prediction mode is set to 6 for fair comparison. 6 layers of Transformer are stacked for the encoder module. The dimension of the hidden feature is set to 256, and the number of heads in all MHA is 8. The learning rate is set to $2.5\times10^{-3}$ initially and decreases over the training epochs. The AdamW optimizer is adopted with a weight decay of $1\times10^{-6}$. We train the model for 45 epochs with a batch size of 24 on a server with 4 NVIDIA 4090s.
\vspace{-0.1cm}

\subsection{Compared Methods}

We consider the existing methods on the V2X-Seq dataset and a baselines we proposed for performance comparison.

\begin{itemize}
\item \textbf{PP-VIC} stitch the AV and infrastructure trajectory through CBMOT, and then test it with the SOAT prediction models TNT \cite{zhao2021tnt} and HiVT \cite{zhou2022hivt}.
\item \textbf{V2X-Graph} constructs a graph to fuse the history trajectories from AV and infrastructure, however, it just focuses on the history domain fusion.
\item \textbf{Co-HTTP} is the baseline model, simplified from our Co-MTP model. It uses the CBMOT \cite{benbarka2021score} method to stitch history trajectories from AV and infrastructure. 

\end{itemize}

\begin{table}[t]
\caption{Performance comparison on the V2X-Seq dataset.}
\label{table: performance}
\renewcommand{\arraystretch}{1.25}
\begin{center}
\begin{tabular}{ccc||ccc}
\hline
\multirow{2}{*}{Model} & \multirow{2}{*}{\shortstack{History\\[0.25ex]Fusion}} & \multirow{2}{*}{\shortstack{Future\\[0.25ex]Fusion}} & \multicolumn{3}{c}{Metrics $\downarrow$ (K=6)}\\
\cline{4-6}
        &                &               & minADE  & minFDE & MR\\                        
\hline
\hline
TNT \cite{zhao2021tnt} & stitch & $\times$ & 7.38 & 15.27 & 0.72\\
HiVT \cite{zhou2022hivt} & stitch & $\times$ & 1.25 & 2.33 & 0.35\\
V2X-Graph \cite{ruan2023learning} & learning & $\times$ & 1.17 & 2.03 & 0.29\\
\rowcolor{gray!10}Co-HTTP & stitch & learning & 0.83 & 1.33 & 0.19\\
\rowcolor{gray!25}\textbf{Co-MTP} & \shortstack{learning$^{*}$} & learning & \textbf{0.76} & \textbf{1.15} & \textbf{0.16}\\
\hline
\end{tabular}
\end{center}
\begin{tablenotes}
    \item * First stitches the AV-perceived trajectory with the infrastructure’s history trajectory, then fuses them through learning.
\end{tablenotes}
\vspace{-0.5cm}
\end{table}

\begin{table*}[h]
\caption{Results of Model Ablation Study.}
\vspace{-0.3cm}
\label{table: Ablation}
\renewcommand{\arraystretch}{1.3}
\setlength{\tabcolsep}{3mm}
\begin{center}
\resizebox{\textwidth}{0.9in}{
\begin{tabular}{m{8ex}cccccc|ccc}
\hline
\makecell[c]{Time\\Dimension} & \makecell[c]{Model\\Variation} & \makecell[c]{Vehicle-side\\History} & \makecell[c]{Infrastructure\\History} & \makecell[c]{Fusion\\Method} & \makecell[c]{AV's\\Planning} & \makecell[c]{Other's\\Future} & minADE $\downarrow$ & minFDE $\downarrow$ & MR $\downarrow$ \\
\hline
\multirow{4}{*}{History} & 1 & \checkmark & - & - & - & - & 0.9887 & 1.5996 & 0.2224\\
                         & 2 & - & \checkmark & - & - & - & 0.9375 & 1.5123 & 0.2081\\
                         & 3 & \checkmark & \checkmark & stitch & - & - & 0.9123 & 1.4647 & 0.2019\\
                         & 4 & \checkmark & \checkmark & stitch + full-infra & - & - & 0.8518 & 1.3368 & 0.1828\\
\hline
\hline
\multirow{4}{*}{Future} & 5 & \checkmark & \checkmark & stitch + full-infra & \checkmark & - & 0.8378 & 1.2649 & 0.1720\\
                        & 6 & \checkmark & \checkmark & stitch + full-infra & - & \checkmark & 0.7992 & 1.1991 & 0.1659\\
                        & Co-HTTP & \checkmark & \checkmark & stitch & \checkmark & \checkmark & 0.8258 & 1.3324 & 0.1916\\
                        \rowcolor{gray!25}& \textbf{Co-MTP} & \checkmark & \checkmark & stitch + full-infra & \checkmark & \checkmark & \textbf{0.7647} & \textbf{1.1470} & \textbf{0.1594}\\
\hline
\end{tabular}}
\end{center}
\begin{tablenotes}
    \item *stitch represents the CBMOT method, while full-infra indicates that all infrastructure information is included as nodes in graph for fusion.
\end{tablenotes}
\vspace{-0.5cm}
\end{table*}
In the Table \ref{table: performance}, we can observe that our cooperative framework Co-MTP ranks first across minADE/minFDE/MR in the benchmark of V2X-Seq dataset, with a 35\% reduction in minADE compared to V2X-Graph. Moreover, the performance of our proposed baseline is also noteworthy, which preliminarily indicates that the heterogeneous graph network and cross-temporal fusion strategies are beneficial for prediction in cooperative scenarios.

\vspace{-0.05cm}
\subsection{Ablation Study}

We investigate the effectiveness of multi-view data processing strategies and the decoder. Ablation studies evaluate 
variations of Co-MTP separately in both history and future time dimensions, and the results are presented in Table \ref{table: Ablation}. 

\noindent \textbf{History Fusion.} To address the two key questions of what history information should be fused and how to fuse it, we design 4 models based on three conditions: whether the vehicle-side data is available, whether the infrastructure data is available, and which fusion method to adopt. First, the infrastructure has an elevated and static position and shows a better performance than the single-agent prediction. Moreover, as the history information from multiple views becomes more comprehensive, the prediction error gradually decreases. Furthermore, comparing variants 3 and 4 reveals that the fusion method involving full infrastructure information performs better, which means the learning method can fuse the trajectory feature better from noises, and the redundant information of infrastructure can benefit the fusion.

\noindent \textbf{Future Fusion.} We aim to demonstrate whether future fusion, specifically the driving intentions of the AV and the prediction output of the infrastructure, are beneficial to the prediction of the AV. Similarly, we design variations for Co-MTP and analyze them alongside the baseline Co-HTTP. Overall, all models incorporating future fusion outperform those that only consider a single history time dimension. Specifically, the accuracy of 5 and 6 is lower than that of Co-MTP, indicating that both the intention of the AV and the other's future trajectories are crucial for the AV's prediction task. Considering both AV planning and the other's future trajectory from infrastructure's prediction, the prediction performance is further improved in Co-HTTP and Co-MTP.

\begin{table}[t]
\caption{Noise Analysis}
\vspace{-0.1cm}
\raggedleft
\renewcommand{\arraystretch}{1.3} 
\begin{tabular}{c|c|c|c}
\rowcolor{gray!25}\diagbox{Model}{Metrics}  & minADE $\downarrow$ & minFDE $\downarrow$ & MR $\downarrow$ \\
\hline
        Co-HTTP-nofut & 0.9314 \tiny0.9123 & 1.5387 \tiny1.4647 & 0.2077 \tiny0.2019 \\
        Co-MTP-nofut & 0.8637 \tiny0.8518 & 1.3571 \tiny1.3368 & 0.1847 \tiny0.1828 \\
        Co-MTP & 0.7998 \tiny0.7647  & 1.2293 \tiny1.1470 & 0.1672 \tiny0.1594
 
\end{tabular}
\label{tab:Noise}
\begin{tablenotes}
    \item * Results in small font represent prediction performance without noise.
\end{tablenotes}
\vspace{-0.1cm}
\end{table}

\begin{table}[t]
\caption{Time Delay Assessment}
\vspace{-0.1cm}
\raggedleft
\renewcommand{\arraystretch}{1.3} 
\begin{tabular}{c|c|c|c}
    \rowcolor{gray!25}\diagbox{Model}{Metrics}  & minADE $\downarrow$ & minFDE $\downarrow$ & MR $\downarrow$ \\
\hline
    Co-HTTP-nofut & 0.9203 \tiny0.9123 & 1.5170 \tiny1.4647 & 0.2064 \tiny0.2019 \\
    Co-MTP-nofut & 0.8696 \tiny0.8518 & 1.3584 \tiny1.3368 & 0.1810 \tiny0.1828 \\
    Co-MTP & 0.7753 \tiny0.7647  & 1.1849 \tiny1.1470 & 0.1643 \tiny0.1594
\end{tabular}
\label{tab:Delay}
\begin{tablenotes}
\item * Results in small font represent prediction performance without delay.
\end{tablenotes}
\vspace{-0.4cm}
\end{table}

\vspace{-0.05cm}
\subsection{Robustness Assessment}

\begin{figure}
    \centering
    \subfloat[Following]{\fbox{\includegraphics[width=0.21\textwidth, height=0.19\textwidth]{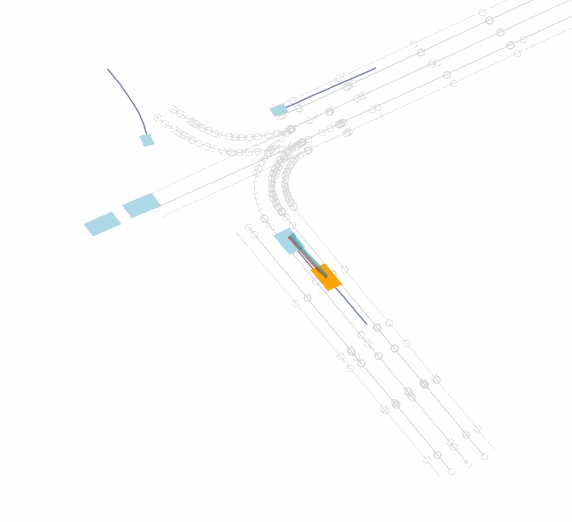}}}
    \hspace{0.5cm}
    \subfloat[Highway]{\fbox{\includegraphics[width=0.21\textwidth, height=0.19\textwidth]{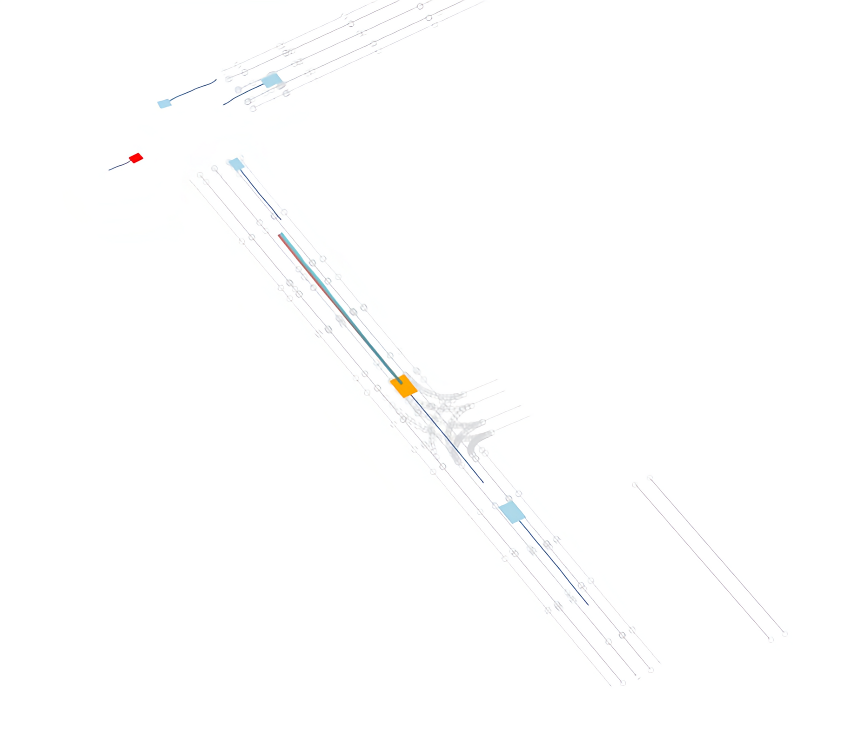}}}
    \hspace{0.5cm}
    \subfloat[Speed up]{\fbox{\includegraphics[width=0.21\textwidth, height=0.19\textwidth]{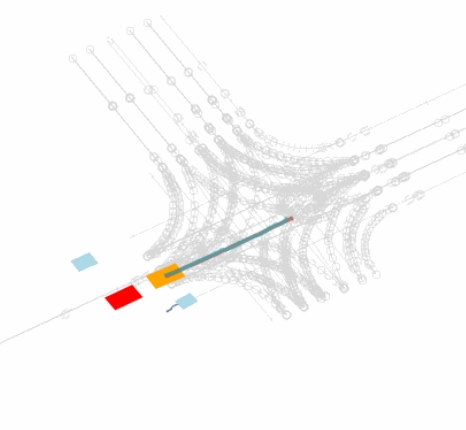}}}
    \hspace{0.5cm}
    \subfloat[Wait to turn]{\fbox{\includegraphics[width=0.21\textwidth, height=0.19\textwidth]{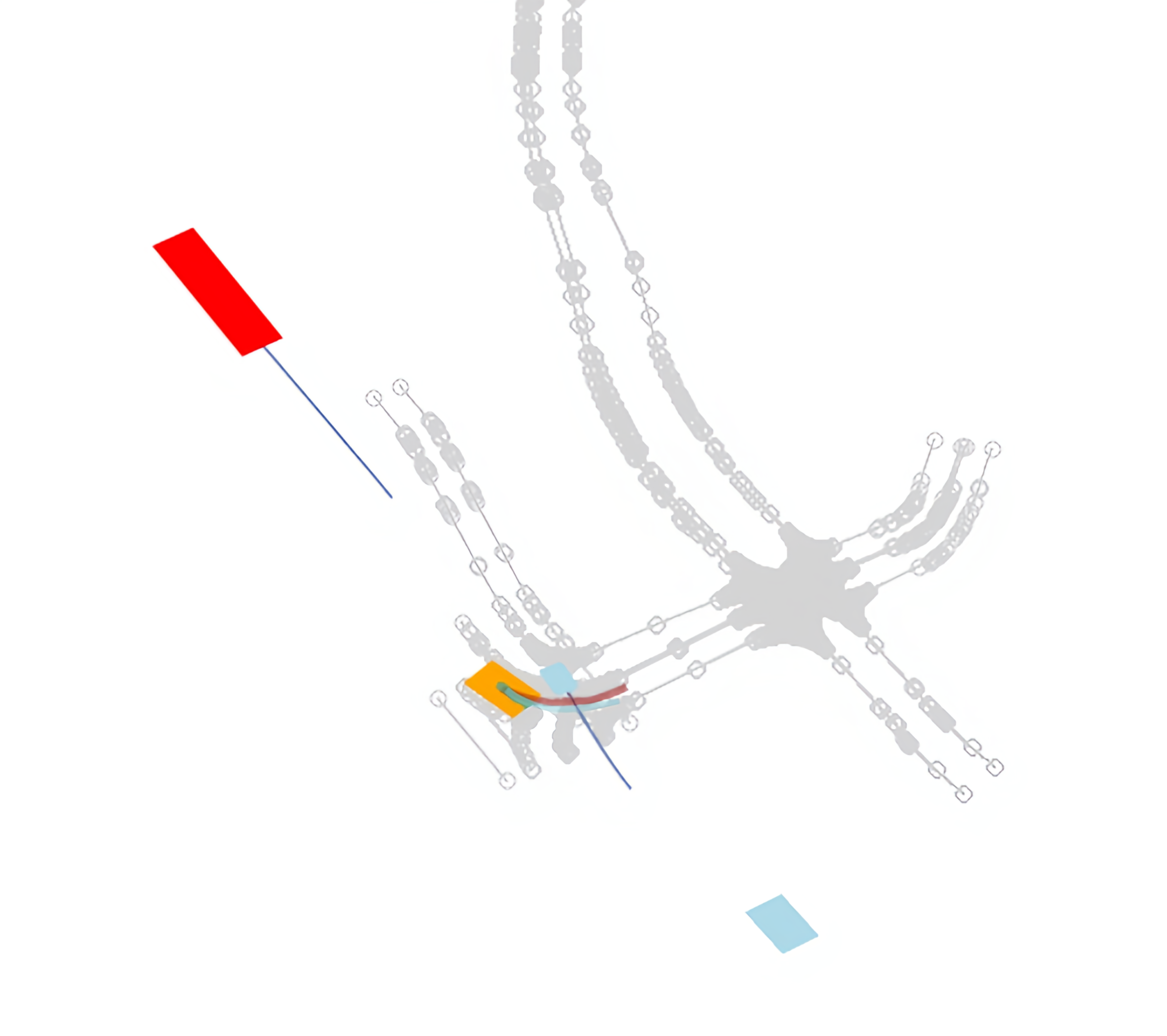}}}
    \caption{Qualitative examples of Co-MTP on V2X-Seq dataset. The red box are AV, while the orange ones are the predicted targets and the blue ones are objects. The predicted trajectories are shown in green, the history ground-truth are shown in blue, and the future ground-truth are shown in brown.}
    \label{fig:subfig_frame}
    \vspace{-0.5cm}
\end{figure}

Co-MTP utilizes V2X communication to fuse both history and future information. While this fusion enriches the temporal understanding, it also introduces common V2X practice challenges such as communication delay and noise. As the amount of communication data increases over the temporal domain, ensuring the model’s robustness against such disruptions becomes more critical. To evaluate this robustness, we designed experiments using the same Co-MTP model base, alongside two variants: Co-MTP-no fusion, which excludes the future fusion, and Co-HTTP-nofut, which simply stitches the trajectory without future information.

We conducted robustness assessments by introducing noise and communication delays, assuming a positional deviation of 0.2 meters and a time delay of 0.5 seconds. The results, presented in Table \ref{tab:Noise} and Table \ref{tab:Delay}, demonstrate that all models—regardless of whether they communicate the history or future data, and whether they use stitching or learning methods—exhibit strong robustness to noise and delay. This indicates that the rich temporal information shared by multiple agents significantly contributes to system robustness, particularly in handling time delays. The robustness to delay is attributed to the system's ability to effectively leverage temporal information, thus mitigating the impact of delay.

In the case of communication noise, the Co-MTP model shows a decline in accuracy. It can be attributed to the compounded errors affecting both history and future domain information, as noise introduces inaccuracies in both. Future research could explore further improvements in noise robustness. Nevertheless, it is noteworthy that despite the performance drop due to noise, Co-MTP continues to demonstrate superior accuracy compared to other models, underscoring its potential in V2X applications.


\vspace{-0.05cm}
\subsection{Qualitative Result}

Figure \ref{fig:subfig_frame} presents the prediction performance of Co-MTP on V2X-Seq in classic driving scenarios, such as following, highway driving, accelerating through an intersection, and waiting to turn. In each sub-figure, the predicted trajectories closely adhere to road structure constraints and traffic safety regulations. Specifically, in (d), the target vehicle intends to turn. However, due to the influence of the object’s intention to go straight without decelerating, the target vehicle is forced to wait. This reinforces the point that future information is crucial for prediction.



\section{CONCLUSION}
To fully exploit comprehensive temporal information in prediction with V2X, we introduce the Co-MTP, the first general cooperative trajectory prediction framework with multi-temporal fusion across both history and future domains. In the history domain, we develop a heterogeneous graph to learn the history fusion from multiple agents with a multi-layer Transformer to handle the incomplete history trajectory issue. To support the future planning of autonomous driving and avoid the overconfidence of the ego vehicle in the planning, we further extended the heterogeneous graph to future domains to capture the future interaction between ego planning and the other vehicles’ actions from infrastructure prediction. Finally, we evaluate the Co-MTP framework on real-world dataset V2X-Seq, and the results show that Co-MTP achieves state-of-the-art performance and that both history and future fusion can greatly benefit prediction. Moreover, we investigate the robustness of V2X and test the Co-MTP performance with delays and noise.

\addtolength{\textheight}{0cm}   








\bibliographystyle{IEEEtran}
\bibliography{root}

\end{document}